\documentclass{article}

\usepackage{PRIMEarxiv}

\usepackage[utf8]{inputenc} 
\usepackage[T1]{fontenc}    
\usepackage{hyperref}       
\usepackage{url}            
\usepackage{booktabs}       
\usepackage{amsfonts}       
\usepackage{nicefrac}       
\usepackage{microtype}      
\usepackage{lipsum}
\usepackage{fancyhdr}       
\usepackage{graphicx}       
\graphicspath{{media/}}     

\usepackage{tikz}
\usepackage{comment}
\usepackage{amsmath,amssymb,amssymb} 
\usepackage{color}

\usepackage{mathrsfs}

\usepackage{marvosym}

\usepackage{multirow}
\newcommand\tvline[1]{ \multirow{#1}{1pt}{$\vcenter{\rule{0.4pt}{\dimexpr#1\baselineskip - 0.2\baselineskip\relax}}$} }

\title{PointAttN: You Only Need Attention for Point Cloud Completion
}

\author{
  Jun Wang, Ying Cui, Dongyan Guo\thanks{guodongyan@zjut.edu.cn} \\
  Zhejiang University of Technology \\
   \And
  Junxia Li, Qingshan Liu \\
  Nanjing University of Information\\
   Science and Technology \\
   \And
  Chunhua Shen \\
  Zhejiang University \\
}

\begin{document}
\maketitle

\begin{abstract}
Point cloud completion referring to completing 3D shapes from partial 3D point clouds is a fundamental problem for 3D point cloud analysis tasks. Benefiting from the development of deep neural networks, researches on point cloud completion have made great progress in recent years. However, the explicit local region partition like kNNs involved in existing methods makes them sensitive to the density distribution of point clouds. Moreover, it serves limited receptive fields that prevent capturing features from long-range context information. To solve the problems, we leverage the cross-attention and self-attention mechanisms to design novel neural network for processing point cloud in a per-point manner to eliminate kNNs. Two essential blocks Geometric Details Perception (GDP) and Self-Feature Augment (SFA) are proposed to establish the short-range and long-range structural relationships directly among points in a simple yet effective way via attention mechanism. Then based on GDP and SFA, we construct a new framework with popular encoder-decoder architecture for point cloud completion. The proposed framework, namely PointAttN, is simple, neat and effective, which can precisely capture the structural information of 3D shapes and predict complete point clouds with highly detailed geometries. Experimental results demonstrate that our PointAttN outperforms state-of-the-art methods by a large margin on popular benchmarks like Completion3D and PCN. 
Code
is available at: \url{https://github.com/ohhhyeahhh/PointAttN}
\end{abstract}


\section{Introduction}
Point cloud completion is the task of estimating a complete shape of an object from its incomplete observation. It plays an important role in 3D computer vision since the raw data captured by existing 3D sensors are usually incomplete and sparse due to factors such as occlusion, limited sensor resolution and light reflection. The unordered and unstructured point cloud data makes the task a challenging problem. Benefiting from the recent advances of deep learning, point cloud completion has achieved remarkable progress. Current popular point cloud completion methods \cite{GAN-RL,pcn,topnet,vrcnet,snowflakenet,PMPNet,RFNet,multi-view,mepcn,Vipc,alliegro2021denoise} mainly revolve around the design of an encoder-decoder architecture for complete point clouds generation. 
The famous point feature extractor PointNet \cite{pointnet} and its variant PointNet++ \cite{pointnet++} are widely taken as the encoders in the mainstream methods \cite{topnet,PFNet,vrcnet,snowflakenet,pointtr,PuNet}. Consequently, the idea of kNN (or its variant ball query) is introduced to build the local spatial relationships among points for local feature learning. The parameter of kNN is fixed and set empirically induced by the density of the point clouds. However, the densities of different point clouds are different, and even within the same point cloud, the points are not uniformly distributed in different local region. Thus, using the fixed parameter to process point clouds is unreasonable, which makes it hard to depict a well local geometric structure of points in local regions. In order to learn the structural features and long-range correlations among local regions, PointTr \cite{pointtr} proposes to adopt Transformers \cite{transformer-is-all-you-need} to build an encoder-decoder architecture. SnowflakeNet \cite{snowflakenet} focuses on the decoding process and introduces skip-transformer to integrate the spatial relationships across different levels of decoding. The attention mechanism in these methods has shown benefits on capturing the structure characteristic in point clouds. However, kNNs are still used in these methods to capture the geometric relation in the point cloud.

\begin{figure}[htbp]
\centerline{\includegraphics[width=1\textwidth]{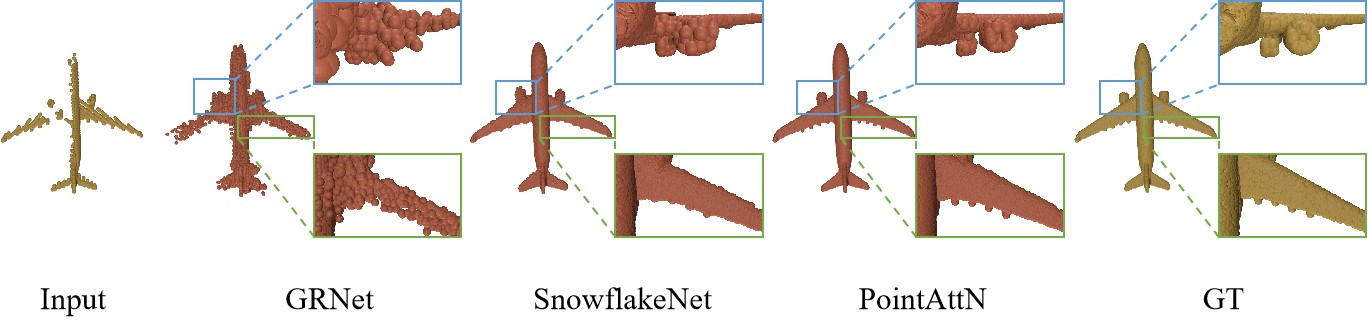}}
\caption{Visual comparisons of the proposed PointAttN and recent methods like SnowflakeNet \cite{snowflakenet} and GR-Net \cite{GRNet}. Our PointAttN can produce high-quality complete shape with much better geometric details, which is almost the same as the ground truth.}
\label{first_page}
\end{figure}

In this paper, we investigate eliminating the above-mentioned issue of kNNs, and show that a properly designed encoder-decoder architecture with only MLP and attention mechanism can achieve leading performance in point cloud completion. Different from existing methods that adopt explicit local feature learner from local neighborhoods to capture local context, the proposed framework, termed PointAttN, focuses on capturing both local and global geometric context in an implicit manner via the self-attention and cross-attention mechanisms. PointAttN adopts the encoder-decoder architecture to process the point completion task in a coarse-to-fine manner. It mainly consists of three parts: a feature extractor module for local geometric structure and global shape feature capturing, a seed generator module for coarse point cloud generation, and a point generator module to produce the fine-grained point cloud. To construct the three modules, we specifically design two essential blocks, one is a geometric details perception block (GDP) and another is a self-feature augment block (SFA). 
The GDP block is designed to adaptively capture the local structural information of the point cloud. It adopts the cross-attention to establish the spatial relationships of points between the original point cloud feature and the down-sampled point cloud feature. By eliminating the concept of explicit local feature region partition like kNNs, GDP can adaptively capture the local shape structure of the point cloud, which can alleviate the influence of local point density and achieves more precise geometric details for reconstructing fine-grained point information. 
The SFA block is designed to reveal detailed geometry of 3D shapes for predicting the complete point cloud. It establishes the spatial relationship among points by introducing self-attention. By cascading SFAs in the coarse-to-fine decoding process, we can progressively reveal the spatial structural and context information of the 3D shape in each processing step to produce more detailed structure. Considering its ability of capturing global information, SFA can also be adopted in the encoding step, which is used to enhance the global structure perception ability of the feature model.
With such designs, the proposed framework PointAttN is simple and neat, which only includes the down-sampling operation of farthest point sampling (FPS), multilayer perceptron (MLP) and attention layers. 

PointAttN eliminates the widely used explicit local region partition for feature learning, which is sensitive to the density distribution of point clouds. By leveraging the cross-attention and self-attention mechanisms, PointAttN can precisely learn both short-range and long-range structural correlations among points, which is crucial for recovering fine-grained information in the decoding phase. Experimental results demonstrate that the proposed method achieves leading performance on challenging benchmarks with such a simple and neat architecture. As shown in Figure~\ref{first_page}, our PointAttN can generate fine-grained shapes with precise geometric details. In summary, our main contributions are:
\begin{itemize}
\itemsep 0pt
	\item  We propose a novel framework PointAttN for point cloud completion. As far as we know, PointAttN is the first time to eliminate the explicit local region operations such as kNNs in Transformer-based methods. It alleviates the influence of the data density distribution and achieves high-quality complete shapes with precise geometrical details retainment.

    \item  We propose two essential blocks GDP and SFA for constructing the framework. They establish the spatial relationships among points in a very simple yet effective way via attention mechanism. Moreover, the blocks can be easily incorporated into other networks to enhance the feature representation capability in large scale point cloud analysis tasks. 
    
    \item  Without bells and whistles, the proposed method achieves the state-of-the-art performance for point cloud completion on challenging benchmarks such as Completion3D \cite{topnet} and PCN \cite{pcn}.
\end{itemize}

\section{Related Work}
Point cloud completion aims to predict the complete shape of an 3D object from the raw partial point cloud. Reconstructing the geometric structure of the point cloud is the key issue for the completion task. Traditional point cloud completion methods usually reconstruct the complete geometry with respect to a predefined smooth surface \cite{Local-frequency,surface-reconstruction}, or build the structural local priors for partial shapes from a large collection of structured basic 3D shapes \cite{Structure-recovery,data-driven}. These methods rely on the prior structured data distribution, which is hard to cover enough cases.

Benefiting from the development of deep neural networks, the geometric information of point clouds can be expressed in high dimensional features, thus the complete shape can be decoded from features of partial point clouds. A new challenge is to capture the topological details of the unordered point clouds. Pioneering works \cite{Shape-completion,High-resolution,PointGrid,Learning-3D} map the point cloud to a voxel grid, and then use 3D convolution to extract features. GRNet \cite{GRNet} further proposes a gridding reverse module to map voxels to point clouds and complete the point cloud in voxel mesh. VE-PCN \cite{VE-PCN} adds shape edges to the voxel mesh by training an edge encoder branch. However, these 3DCNNs based methods suffer from expensive memory and computational cost. Moreover, due to the cubic nature of the voxel mesh, features of the point cloud surface cannot be properly represented. 

With the success of PointNet \cite{pointnet} that directly processes 3D coordinates, many researchers leverage it as the feature encoder and pay special attention on the decoding process to produce complete point clouds. TopNet \cite{topnet} proposes to enhance the decoder with a rooted tree structure, which reconstructs the topology of the point clouds. PFNet \cite{PFNet} proposes to get multi-scale point cloud inputs and fuse different fine-grained features to generate the complete shape.
However, since PointNet \cite{pointnet} directly processes all the points with max pooling to obtain global features, the local structures among points are not learned by the network, which lead to the loss of shape details during decoding. To solve the problem, Zhang et. al. \cite{NSFA} proposes to explore the functionality of multi-scale features from different layers to enhance the performance. CDN \cite{cascade} proposes a cascaded refinement network to synthesize the detailed object shapes by considering both the local details of partial input with the global shape information together. SpareNet \cite{style-based} presents the channel-attentive EdgeConv to exploit the local structures and the global shape in point features.
Inspired by the different receptive fields across multiple levels of CNNs, PointNet++ \cite{pointnet++} proposes to process a set of points sampled in a metric space in a hierarchical fashion, where the ball query (an invariant of kNN) is used to guarantee local neighborhoods. By leveraging its success representation of local shape features, recent works with encoder-decoder architecture that adopt PointNet++ to construct feature extractors have shown great progress in point cloud completion \cite{vrcnet,SA-Net,snowflakenet}. However, the prefixed partition of local regions involved by kNNs make these methods sensitive to the density of point clouds. Moreover, the involved limited receptive fields prevent the feature extractor from achieving better local and global structural information of the point cloud.

Compared with the limited receptive fields of CNNs, Transformer \cite{transformer-is-all-you-need} characterized by the attention mechanism shows its advantages in long-range interaction capturing. In recent years, Transformer-based methods have gradually replaced CNNs and show their strong capability in many computer vision fields \cite{NLN,swim,vit,detr,setr,SiamGAT,TransT}. Inspired by such success, researchers try to introduce the transformer framework into point cloud analysis tasks \cite{PCT,pointformer,SA-Net,point-transformer}.
PointTransformer \cite{point-transformer} designs a Point Transformer layer for point cloud processing. SA-Net \cite{SA-Net} introduces the skip-attention mechanism to fuse local region features from encoder into point features of decoder, which enables more detailed geometry information preserving for decoding. On the other hand, SnowflakeNet\cite{snowflakenet} and PointTr\cite{pointtr} pay special attention on the decoding process via Transformer architectures. These methods have demonstrated the capability of transformers in point cloud completion. However, due to the memory constraints or explicit local part demand, kNNs are still utilized in these methods.
In this work, we show that kNNs can be eliminated by a properly designed framework which fully exploits the advantages of self-attention and cross-attention mechanisms. Meanwhile, we demonstrate that a much simpler architecture can achieve even better performance than state-of-the-art methods for point cloud completion.

\section{Proposed Method}

\begin{figure*}[htbp]
\centerline{\includegraphics[width=1\textwidth]{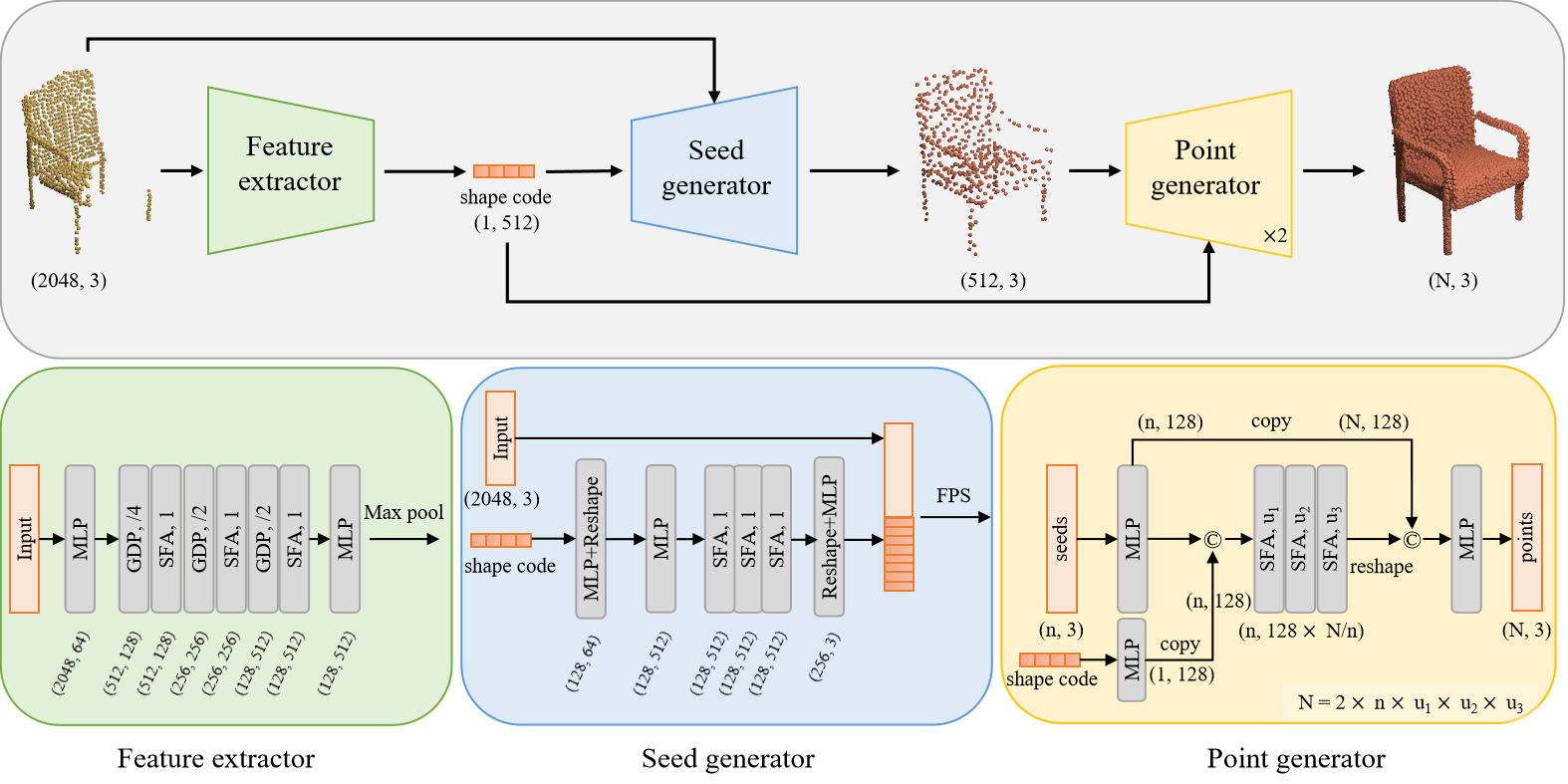}}
\caption{Illustration of PointAttN. The proposed framework consists of three modules: the feature extractor, seed generator and point generator. The overall architecture of PointAttN is shown in the top, and the details of the three modules are show in the bottom. Here $\copyright$ denotes a concatenation operation. It should be noticed that all the MLPs are applied on per-point features. For easy understanding, we take 2048 as the number of input partial points to draw the illustration. Point clouds with other number of partial points can also be processed by retraining the model.}
\label{PointAttn}
\end{figure*}

The overall framework of the proposed PointAttN is illustrated in Figure~\ref{PointAttn}. The framework adopts the popular encoder-decoder architecture for point cloud completion. It mainly consists of three modules, a feature extractor for shape feature encoding, a seed generator module and a point generator module for coarse-to-fine generation of the complete shape. The three modules are designed upon 
the proposed Geometric Details Perception (GDP) and Self-Feature Augment (SFA), which are two essential blocks for processing the point cloud data. GDP and SFA are drawn on the core idea of Transformer \cite{transformer-is-all-you-need}, i.e., introducing the multi-head cross-attention and self-attention mechanisms. However, the encoder-decoder structure of Transformer is not adopted in our method.
In the following, we will first introduce the two essential blocks and then describe the three modules based on them in details.

\subsection{Geometric Details Perception}
Capturing the local geometric information is crucial but challenging for point cloud completion. Existing local shape models with kNNs are easily affected by local density of the point cloud. To tackle the problem, we design a geometric details perception (GDP) block by leveraging the cross-attention to adaptively aggregate information from unordered points, which can implicitly model local features of the point cloud. 

\begin{figure}[htbp]
\centerline{\includegraphics[width=0.80\textwidth]{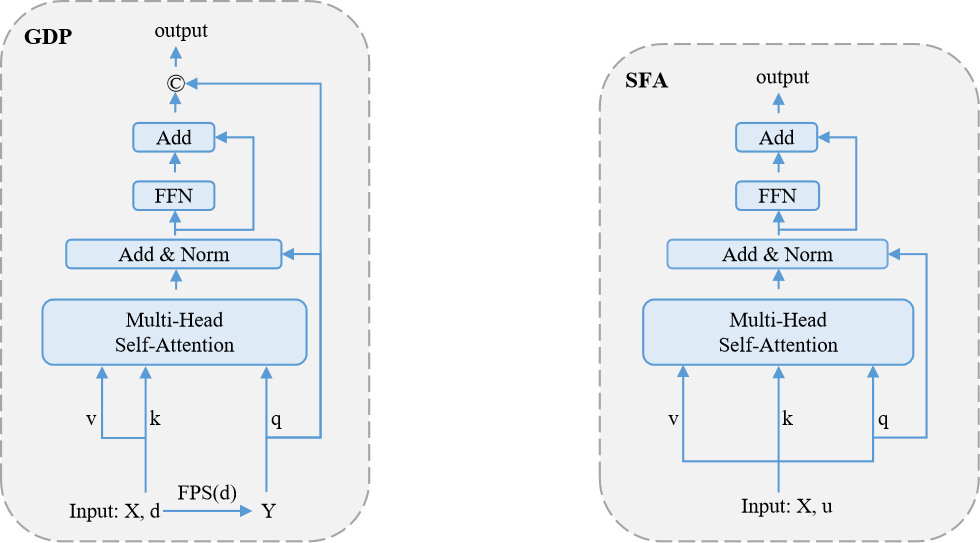}}
\caption{The detailed structure of GDP (left) and SFA (right). Here $\copyright$ denotes a concatenation operation.}
\label{gdp-and-sfa}
\end{figure}

The structure of GDP is illustrated in the left of Figure~\ref{gdp-and-sfa}. GDP receives an input $(X, d)$, where $X$ is a matrix of size $n \times c$ and each row of $X$ can be considered as a feature vector corresponding to a point, $d$ is the down-sampling ratio. By applying a farthest point sampling (FPS) algorithm \cite{pointnet++}, we can get a down-sampled point cloud feature matrix $Y$ of size $n/d \times c$. Then we leverage the multi-head cross-atttention in the form of residual to learn the feature matrix $F$:
\begin{equation}
\begin{split}
    F=Norm(Q+MultiHead(Q,K,V)),\\
    Q=YW^{Q}, K=XW^{KV}, V=XW^{KV},
\end{split}
\end{equation}
where $MultiHead(\cdot)$ is performed similarly to \cite{transformer-is-all-you-need}, $W^{Q}\in{\mathbb{R}^{c\times c}}$ and $W^{KV}\in{\mathbb{R}^{c\times c}}$ are linear transformation matrices.
Through this operation, each point in $Y$ can adaptively aggregate similar point features (including similar shape) from $X$, where the similarity may include similarity in shape and proximity in distance, thus the local geometric structure can be perceived in the model.

In order to enhance the fitting ability of the model, we utilize a feed forward network (FFN) \cite{transformer-is-all-you-need} to further update $F$. The final output of GDP can be formulated as
\begin{equation}
GDP(X,d)=Concat(F+FFN(X), Q).
\end{equation}
\label{eq_gdp}
where $Concat(\cdot)$ denotes a concatenation operation.

It should be noticed that all the involved linear transformations in GDP do not change the feature dimension. The output of GDP is a matrix of size $(n/d)\times 2c$.

\subsection{Self-Feature Augment}
The self-feature augment (SFA) is proposed to reveal detailed geometry of 3D shapes for predicting the complete point cloud. It establishes the spatial relationship among points by introducing self-attention. Considering its ability of capturing global information, SFA can be also adopted for feature extraction to ehance the feature representation ability.

The structure of SFA is illustrated in the right of Figure~\ref{gdp-and-sfa}.
SFA receives an input $(X, u)$, where $X$ is a matrix of $n \times c$, $u$ is an up-sampling ratio. SFA integrates the information from different points of $X$ by applying the multi-head self-attention in the form of residual:
\begin{equation}
\begin{split}
    F=Norm(Q+MultiHead(Q,K,V)),\\
    Q=XW^{Qu},K=XW^{KVu},V=XW^{KVu},
\end{split}
\end{equation}
where $W^{Qu}\in{\mathbb{R}^{c\times uc}}$ and $W^{KVu}\in{\mathbb{R}^{c\times uc}}$ are linear transformation matrices. The dimension of point features is increased by $u$ times after the linear transformation operation.

Similar to GDP, a FFN is also ultilized here, thus the output of SFA can be computed as
\begin{equation}
SFA(X,u)=F+FFN(X).
\end{equation}
Accordingly, the output of SFA is a matrix of size $n\times uc$.

\subsection{PointAttN for Point Cloud Completion}
The proposed framework PointAttN adopts the popular encoder-decoder architecture for point cloud completion. The encoder module as Feature Extractor is constructed upon both GDP and SFA blocks to capture precise geometric information of 3D shapes. Meanwhile, the decoder modules including Seed Generator and Point Generator are constructed upon cascade SFAs to produce complete point cloud in a coarse-to-fine decoding manner.

\textbf{Feature Extractor.}
The proposed framework PointAttN takes the partial point cloud as input. 
The feature extractor aims to produce a shape code that captures both local structural details and global context of the object for the following coarse-to-fine shape decoding.
As shown in the lower left of Figure~\ref{PointAttn}, we first adopt multi-layer perceptron (MLP) to map the unordered points into feature vectors. Then we alternately stack three GDP blocks and three SFA blocks, through which the local and global information can be progressively embedded during the down-sampling process. GDP is introduced to perceive the local structure of point clouds. The down-sampling ratio of the three GDP blocks is set as $4, 2, 2$, respectively. Meanwhile, SFA is adopt to enhance the global information perception of the model after each GDP operation. Here the up-sampling ratio of the three SFA blocks is set as 1, that is, SFA in this module do not change the dimension of features. After the stacks, another MLP is adopt as the role of FFN mentioned above and a max pooling operation is utilized to generate the shape code as output of this module.

\textbf{Seed Generator.}
Based on decoding the input shape code, the seed generator aims to produce a sparse but complete point cloud, which retains the geometric structure of the target complete shape. 
As shown in the lower middle of Figure~\ref{PointAttn}, the shape code is first decoded into a feature matrix through MLP and a reshape operation, where each row of the matrix can be considered as a feature vector of a point. After extending the dimension of point features through another MLP, three SFA blocks are cascaded to enhance the feature ability of perceiving the target geometric structure. Next, the sparse points are produced by splitting the point features through reshape and MLP, where MLP is used to transform the corresponding feature vectors into 3D position coordinates. Then, the sparse points are merged with the input partial point cloud by concatenation. The merged point cloud is then down-sampled by FPS to produce the sparse point cloud, which is served as the seed point cloud for the next point generator module.

\textbf{Point Generator.}
The point generator module receives both the shape code and the sparse but complete point cloud as input and output the fine-grained point cloud.
As shown in the lower right of Figure~\ref{PointAttn}, the seed point cloud and the shape code are sent to MLP respectively to generate two point feature matrices $F_1$ and $F_2$ (expanded to the same number of rows with $F_1$ by self-copy), which are then integrated through point-wise concatenation. By that, each point feature in the integrated matrix maintains both local details and global geometric structure of the shape.
In the next, three SFA blocks are cascaded to progressively up-sample the point features, which are then split by a reshape operation. The split point features are then integrated with $F_1$ (expanded to the same number of rows with the split point features by self-copy) through point-wise concatenation. At last, the dense points are produced by applying MLP to the integrated features, where MLP is used to transform the corresponding feature vectors into 3D position coordinates.
During implement, two such point generator modules are cascaded to construct the fine decoding network. Here the up-sampling ratios $u_1, u_2, u_3$ corresponding to the three SFAs are intuitively set without careful tuning, thus the performance of the model may still be further improved by grid search.

It can be observed that our framework is designed without any complicated operations. Such a simple network that achieving leading performance for point cloud completion has fully demonstrated the capability of the proposed GDP and SFA blocks. Meanwhile, the blocks can be easily incorporated into other networks for large scale point cloud analysis tasks to enhance the feature ability of local geometric details and global context perception.

\subsection{Training Loss}
During training, we utilize Chamfer distance (CD) as the metric distance of point cloud to formulate the loss function. Suppose the seed point cloud, the output point clouds of the two cascaded point generators are denoted as $\mathcal{P}_0$, $\mathcal{P}_1$, $\mathcal{P}_2$, respectively. Meanwhile, the groud-truth point cloud is down-sampled through FPS to obtain three sub-clouds $\mathcal{S}_0$, $\mathcal{S}_1$, $\mathcal{S}_2$, which respectively share the same density of $\mathcal{P}_0$, $\mathcal{P}_1$ and $\mathcal{P}_2$.
Then the loss of the model can be defined as follows:

\begin{equation}
\mathcal{L} = \sum_{i=0}^{2}\lambda_{i}d_{CD}(\mathcal{P}_i,\mathcal{S}_i)
\end{equation}
where $d_{CD}$ is the Chamfer distance loss, which is defined as
\begin{equation}
    d_{CD}(\mathcal{P},\mathcal{S})=\frac{1}{\left | \mathcal{P} \right |}\sum_{p\in \mathcal{P}}\underset{s\in \mathcal{S}}{min}{\left \| p-s \right \|} + \frac{1}{\left | \mathcal{S} \right |}\sum_{s\in \mathcal{S}}\underset{p\in \mathcal{P}}{min}{\left \| s-p \right \|}
\end{equation}
In the implementation, each $\lambda_i$ is set as 1.


\section{Experiments}
To validate the effectiveness of our PointAttN, we conduct comprehensive experiments on two challenging benchmarks, namely Completion3D \cite{topnet} and PCN \cite{pcn}, both of which are derived from the ShapeNet \cite{shapenet} dataset.

\textbf{The Completion3D Dataset.}
The Completion3D dataset contains 30958 models of 8 categories, in which both the partial and ground truth point clouds have the same size of $2048 \times 3$. A specific train/validation/test split is given by Completion3D for a fair comparison among point cloud completion methods, where 28974 models, 800 and 1184 models are included in the training set, validation and testing set, respectively. To align with previous works, we use the specified training set to train the model and adopt the L2 version of Chamfer distance (CD) to evaluate the results during experiments.

\textbf{The PCN Dataset.}
PCN is a widely used point cloud completion benchmark that contains 30974 models of 8 categories. The partial point clouds are generated by back-projecting the 3D shape into 8 different partial views and each point cloud has less than 2048 points. Each ground-truth point cloud contains 16384 points, which are evenly sampled from the shape surface. For a fair comparison, we follow the same split settings with PCN\cite{pcn} during experiments and adopt the L1 version of Chamfer distance to evaluate our results.

\textbf{Implementation Details.}
The proposed framework is implemented in Python with PyTorch and trained on 4 NVIDIA 2080Ti GPUs. Models are trained with Adam optimizer by totally 400 epochs, while the learning rate is initialized to 1E-4 and decayed by 0.7 every 40 epochs. The batch size is set to 32. For experiments on the Completion3D dataset, the up-sampling ratios $(u_1, u_2, u_3)$ of the two cascaded point generators are set as $(1, 1, 1)$ and $(1, 1, 1)$, respectively. For experiments on the PCN dataset, the corresponding values are set as $(2, 1, 1)$ and $(2, 1, 2)$, respectively.

\begin{table*}[htbp]
\caption{Point cloud completion results on the Completion3D dataset in terms of L2 Chamfer distance (CD). The best results are highlighted in bold.}
\begin{center}
\begin{tabular}{ l c c c c c c c c c c c }
 \hline
 Methods & \tvline{1} & Average & \tvline{1} & Plane & Cabinet & Car & Chair & Lamp & Couch & Table & Boat \\
 \hline
 FoldingNet\cite{foldingNet} & \tvline{10} &  19.07 & \tvline{10} & 12.83 & 23.01 & 14.88 & 25.69 & 21.79 & 21.31 & 20.71 & 11.51 \\
 PCN\cite{pcn} & & 18.22 & & 9.79 & 22.70 & 12.43 & 25.14 & 22.72 & 20.26 & 20.27 & 11.73 \\
 AtlasNet\cite{atlas} & & 17.77 & & 10.36 & 23.40 & 13.40 & 24.16 & 20.24 & 20.82 & 17.52 & 11.62 \\
 SoftPoolNet\cite{SoftPoolNet} & & 16.15 & & 5.81 & 24.53 & 11.35 & 23.63 & 18.54 & 20.34 & 16.89 & 7.14 \\
 TopNet\cite{topnet} & & 14.25 & & 7.32 & 18.77 & 12.88 & 19.82 & 14.60 & 16.29 & 14.89 & 8.82 \\
 SA-Net\cite{SA-Net} & & 11.22 & & 5.27 & 14.45 & 7.78 & 13.67 & 13.53 & 14.22 & 11.75 & 8.84 \\
 GRNet\cite{GRNet} & & 10.64 & & 6.13 & 16.90 & 8.27 & 12.23 & 10.22 & 14.93 & 10.08 & 5.86 \\
 PMP-Net\cite{PMPNet} & & 9.23 & & 3.99 & 14.70 & 8.55 & 10.21 & 9.27 & 12.43 & 8.51 & 5.77 \\
 VRC-Net\cite{vrcnet} & & 8.12 & & 3.94 & 10.93 & 6.44 & 9.32 & 8.32 & 11.35 & 8.60 & 5.78 \\
 SnowflakeNet\cite{snowflakenet} & & 7.60 & & 3.48 & 11.09 & 6.90 & 8.75 & 8.42 & 10.15 & 6.46 & 5.32 \\
 \hline
 ours & \tvline{1} & \textbf{6.63} & \tvline{1} & \textbf{3.28} & \textbf{10.77} & \textbf{6.13} & \textbf{7.14} & \textbf{5.92} & \textbf{9.72} & \textbf{6.16} & \textbf{3.59} \\
 \hline
\end{tabular}
\end{center}
\label{table-C3D}
\end{table*}

\begin{figure*}[htbp]
\centerline{\includegraphics[width=1\textwidth]{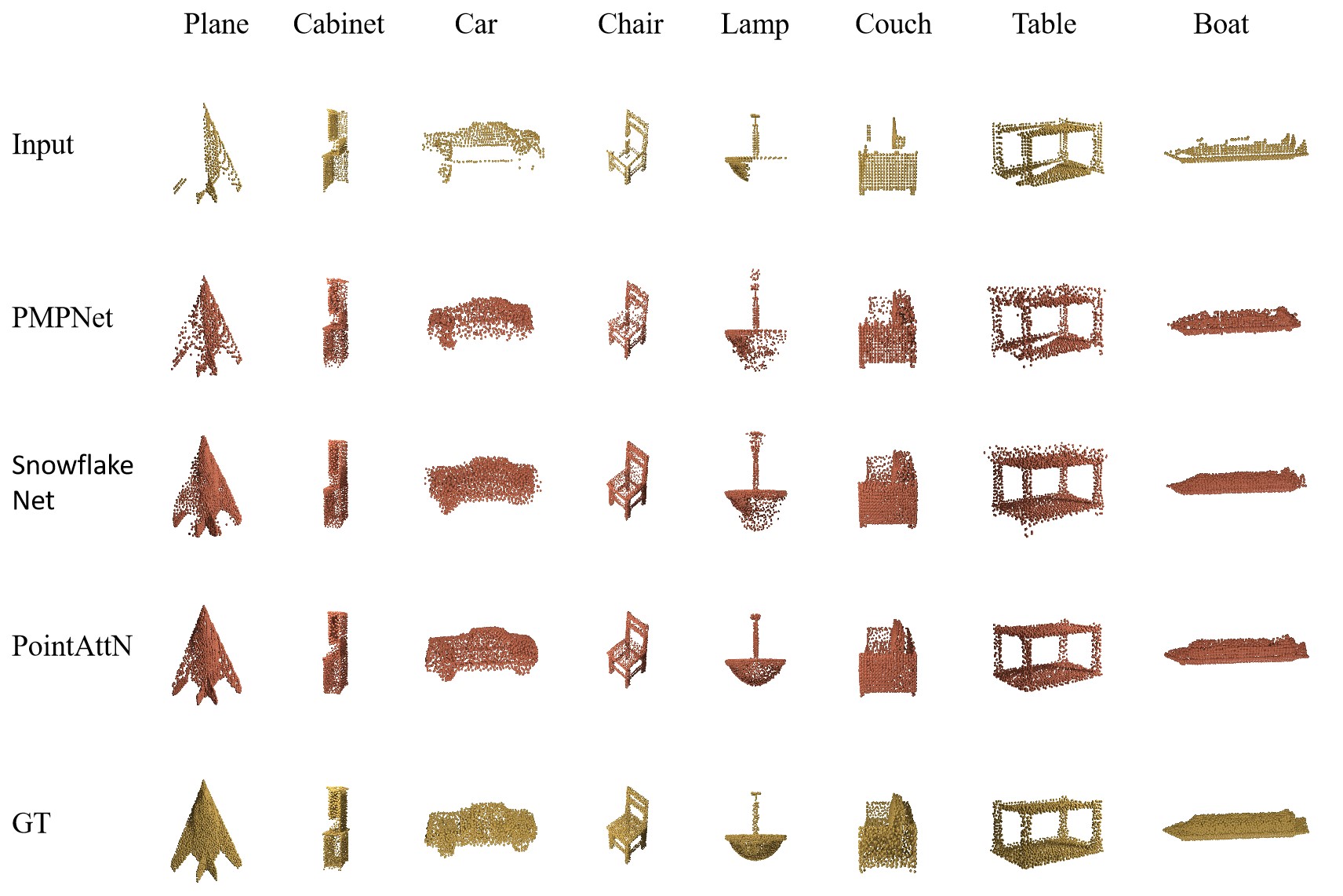}}
\caption{Visual comparisons of point cloud completion on the Completion3D dataset. Compared with other state-of-the-art methods, our PointAttN can produce much more fine-grained geometric structure of the shapes (e.g., lamp chair and boat) and much smoother surfaces (e.g., car and couch).}
\label{visualc3d}
\end{figure*}

\begin{table*}[htbp]
\caption{Point cloud completion results on the PCN dataset in terms of L1 Chamfer distance (CD). The best results are highlighted in bold.}
\begin{center}
\begin{tabular}{ l c c c c c c c c c c c }
 \hline
 Methods & \tvline{1} & Average & \tvline{1} & Plane& Cabinet & Car & Chair & Lamp & Couch & Table & Boat \\
 \hline
 FoldingNet\cite{foldingNet} & \tvline{11} & 14.31 & \tvline{11} & 9.49 & 15.80 & 12.61 & 15.55 & 16.41 & 15.97 & 13.65 & 14.99 \\
 AtlasNet\cite{atlas} &  & 10.85 &  & 6.37 & 11.94 & 10.10 & 12.06 & 12.37 & 12.99 & 10.33 & 10.61 \\
 PCN\cite{pcn} &  & 9.64 &  & 5.50 & 22.70 & 10.63 & 8.70 & 11.00 & 11.34 & 11.68 & 8.59 \\
 TopNet\cite{topnet} &  & 12.15 &  & 7.61 & 13.31 & 10.90 & 13.82 & 14.44 & 14.78 & 11.22 & 11.12 \\
 GRNet\cite{GRNet} &  & 8.83 &  & 6.45 & 10.37 & 9.45 & 9.41 & 7.96 & 10.51 & 8.44 & 8.04 \\
 PMP-Net\cite{PMPNet} &  & 8.73 &  & 5.65 & 11.24 & 9.64 & 9.51 & 6.95 & 10.83 & 8.72 & 7.25 \\
 CRN\cite{cascade} &  & 8.51 &  & 4.79 & 9.97 & 8.31 & 9.49 & 8.94 & 10.69 & 7.81 & 8.05 \\
 PointTr\cite{pointtr} &  & 8.38 &  & 4.75 & 10.47 & 8.68 & 9.39 & 7.75 & 10.93 & 7.78 & 7.29 \\
 NSFA\cite{NSFA} &  & 8.06 &  & 4.76 & 10.18 & 8.63 & 8.53 & 7.03 & 10.53 & 7.35 & 7.48 \\
 SA-Net\cite{SA-Net} &  & 7.74 &  & \textbf{2.18} & 9.11 & \textbf{5.56} & 8.94 & 9.98 & \textbf{7.83} & 9.94 & 7.24 \\
 SnowflakeNet\cite{snowflakenet} &  & 7.21 &  & 4.29 & 9.16 & 8.08 & 7.89 & 6.07 & 9.23 & 6.55 & 6.40 \\
 \hline
 ours & \tvline{1} & \textbf{6.86} & \tvline{1} & 3.87 & \textbf{9.00} & 7.63 & \textbf{7.43} & \textbf{5.90} & 8.68 & \textbf{6.32} & \textbf{6.09} \\
 \hline
\end{tabular}
\end{center}
\label{table-pcn}
\end{table*}

\begin{figure*}[htbp]
\centerline{\includegraphics[width=1\textwidth]{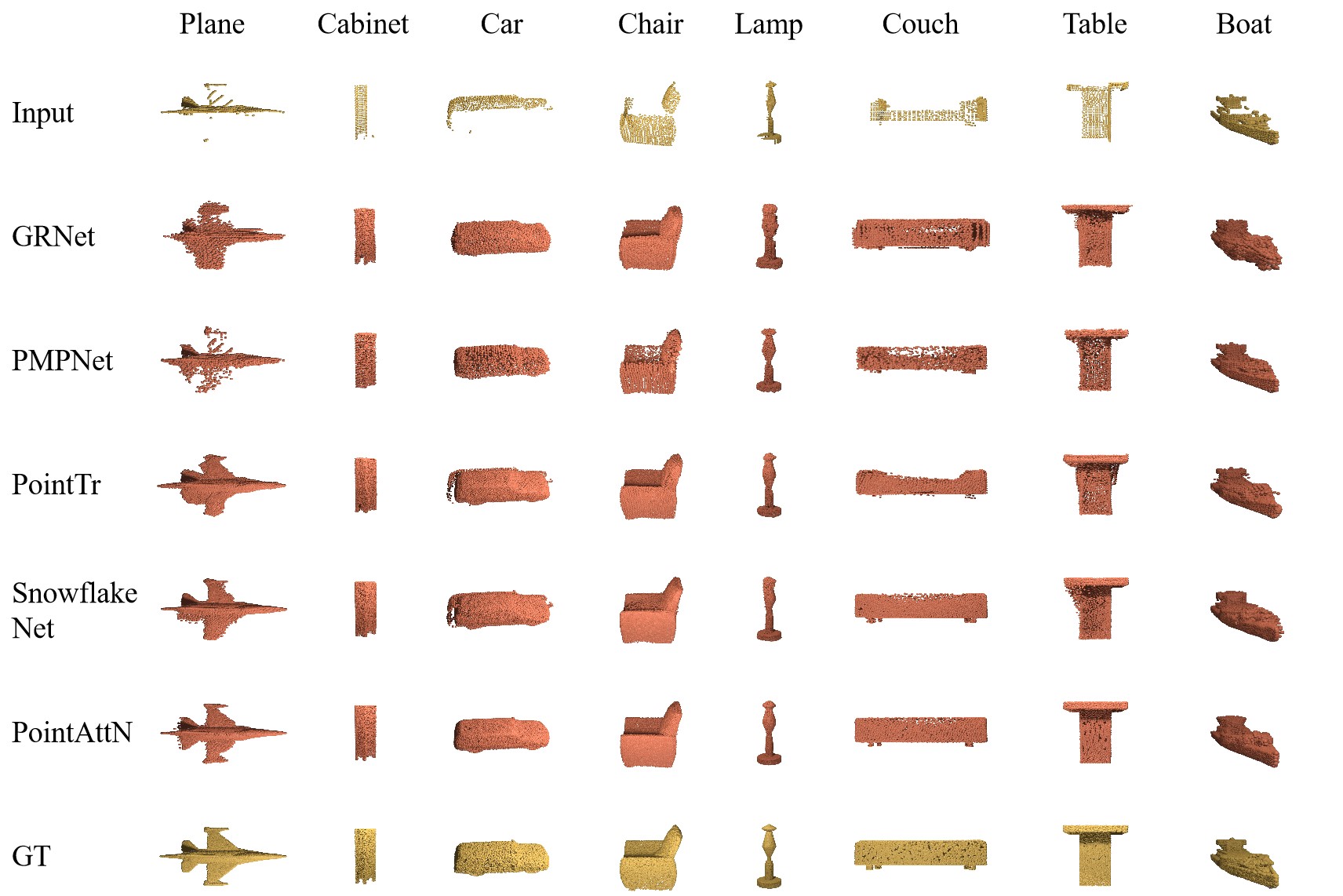}}
\caption{Visual comparisons of point cloud completion on the PCN dataset. Compared with other state-of-the-art methods, our PointAttN can produce much more fine-grained geometric structure of the shapes (e.g., lamp, boat and plane) and much smoother surfaces (e.g., table, car, chair and couch).}
\label{visualpcn}
\end{figure*}

\subsection{Results on Completion3D}
We evaluate PointAttN on the Completion3D dataset and compare it with state-of-the-art methods. The comparison results are shown in Table~\ref{table-C3D}, where all results of the methods are cited from the public leaderboard of Completion3D benchmark.
Table~\ref{table-C3D} shows that our PointAttN ranks the first on all categories in terms of CD, which demonstrate the generalization capability of PointAttN. 
Compared with the second-ranked method SnowflakeNet \cite{snowflakenet}, our PointAttN reduces the average CD by 0.97, which is 12.8\% lower than that of SnowflakeNet (6.63 versus 7.60).
Especially, for the boat category our PointAttN achieve a reduction of 32.5\% compared with the second-ranked SnowflakeNet \cite{snowflakenet} in terms of per-category CD (3.59 versus 5.32), while for the lamp category our PointAttN achieve a reduction of 28.8\% compared with the second-ranked VRC-Net \cite{vrcnet} (5.92 versus 8.32).
Compared with the methods like VRC-Net \cite{vrcnet} and SnowflakeNet \cite{snowflakenet} that also adopt the same coarse-to-fine decoding strategy, our PointAttN achieves significant improvement, which owing to the proposed GDP and SFA blocks in the framework that help to capture both local and global structure information of the shape.

A visual comparison on all 8 categories is shown in Figure~\ref{visualc3d}. Obviously, our PointAttN can generate much better complete point clouds with high shape quality.
For example, our method can generate much smoother surfaces than other methods for the plane, car, chair and table categories, while other methods generate noisy points or uneven point distribution. And for the chair, lamp, couch and boat categories, PointAttN can achieve more detailed and precise geometries of the 3D shapes, especially on the lampshade and the complex boat shape.

\subsection{Results on PCN}
The quantitative results of our PointAttN and state-of-the-art methods on the PCN dataset are shown in Table~\ref{table-pcn}, where the results of other methods are cited from their papers. 
It can be seen from Table~\ref{table-pcn} that our method achieves an improvement of 4.9\% reduction in terms of average CD compared with the second-ranked SnowflakeNet \cite{snowflakenet} (6.86 versus 7.21). 
Especially, compared with the state-of-the-art Transformer-based methods like SnowflakeNet \cite{snowflakenet}, SA-Net \cite{SA-Net} and PointTr \cite{pointtr}, our PointAttN surpasses SnowflakeNet \cite{snowflakenet} and PointTr \cite{pointtr} on average and all categories, while surpasses SA-Net \cite{SA-Net} on average and most of the categories like cabinet, chair, lamp, table and boat. The comparison demonstrate the capability of our attention-based framework without kNNs for producing complete shapes.

To further evaluate the performance, a comparison of visualization results is shown in Figure~\ref{visualpcn}.
From the figure we can see that our PointAttN also achieves much better visual results than other state-of-the-art methods on the PCN dataset, including much smoother surfaces and much more precise geometric structures. Compared with the Transformer-based methods SnowflakeNet \cite{snowflakenet} and PointTr \cite{pointtr} that still adopt explicit kNNs, our PointAttN especially achieves much more precise geometric structures and local details as shown in the plane (geometric structure of the wings), car (details of the trunk) and lamp (geometric structure of the global shape) categories. That is because, our framework without kNN eliminates the limited receptive field of explicit local partition and better perceives both local details and global structures during the process. 

\subsection{Ablation study}
To verify the effectiveness, we construct a detailed ablation study of each part in PointAttN on the Completion3D dataset. 
Five different network variations are designed as follows.

(1) The \textsl {FE~w/~GDP} variation.
To verify the effectiveness of the proposed GDP for feature extraction, we replace all the GDP blocks with FPS that has the same down-sampling function. To maintain the model structure, the up-sampling ratio $u$ of all SFAs in the Feature Extractor (FE) module is set to 2.

(2) The \textsl{FE~w/o~SFA} variation.
To verify the effectiveness of the proposed SFA for feature extraction, we remove all the SFA blocks in the Feature Extractor (FE) module and keep all other structures unchanged.

(3) The \textsl{FE+FoldingNet} variation.
To verify the feature extracting ability of the designed Feature Extractor (FE) module, we use it to replace the backbone of FoldingNet \cite{foldingNet}.

(4) The \textsl{FE+SG+SPD} variation.
To verify the effectiveness of the Feature Extractor (FE) and Seed Generator (SG) modules, we replace the point generator module with the SPD decoder in SnowflakeNet \cite{snowflakenet} and keep other modules unchanged.

(5) The \textsl{PN+SG+PG} variation.
To verify the effectiveness of the Seed Generator (SG) and Point Generator (PG) modules, we replace the Feature Extractor (FE) module with PointNet++ \cite{pointnet++} and keep other modules unchanged. 

\begin{table*}[htbp]
\caption{Ablation study of each part in PointAttN on the Completion3D dataset. Here SA-Net \cite{SA-Net}, FoldingNet \cite{foldingNet} and SnowflakeNet \cite{snowflakenet} are listed as baselines.}
\begin{center}
\begin{tabular}{ l c c c c c c c c c c c }
 \hline
 Evaluate & \tvline{1} & Average & \tvline{1} & Plane& Cabinet & Car & Chair & Lamp & Couch & Table & Boat \\
 \hline
 \textsl{FE~w/o~GDP} & \tvline{8} & 7.42 & \tvline{8} & 2.99 & 13.03 & 6.46 & 7.51 & 6.61 & 11.19 & 7.35 & 3.79 \\
 \textsl{FE~w/o~SFA} &   & 6.81 &   & 3.43 & 11.62 & 6.79 & 7.11 & 6.32 & 9.13 & 6.12 & 3.60 \\
 \textsl{FE+FoldingNet} &   & 8.28 &   & 3.33 & 11.54 & 7.5 & 8.62 & 8.4 & 12.57 & 8.43 & 5.58 \\
 \textsl{FE+SG+SPD} &   & 7.31 &   & 3.19 & 11.49 & 7.26 & 7.78 & 7.13 & 10.71 & 6.48 & 4.08 \\
 \textsl{PN+SG+PG} &   & 7.23 &   & 3.44 & 11.5 & 6.3 & 7.54 & 6.74 & 10.62 & 6.28 & 5.23 \\
  SA-Net\cite{SA-Net} & & 11.22 & & 5.27 & 14.45 & 7.78 & 13.67 & 13.53 & 14.22 & 11.75 & 8.84 \\
   FoldingNet\cite{foldingNet} &  &  19.07 &  & 12.83 & 23.01 & 14.88 & 25.69 & 21.79 & 21.31 & 20.71 & 11.51 \\
   SnowflakeNet\cite{snowflakenet} & & 7.60 & & 3.48 & 11.09 & 6.90 & 8.75 & 8.42 & 10.15 & 6.46 & 5.32 \\
 \hline
 PointAttN & \tvline{1} & 6.63 & \tvline{1} & 3.28 & 10.77 & 6.13 & 7.14 & 5.92 & 9.72 & 6.16 & 3.59 \\
 \hline
\end{tabular}
\end{center}
\label{table-ablation}
\end{table*}

For all the experiments, we only replace corresponding parts and do not change the other experiment settings. The ablation results are shown in Table~\ref{table-ablation} along with the default framework PointAttN and the baselines SA-Net \cite{SA-Net}, FoldingNet \cite{foldingNet} and SnowflakeNet \cite{snowflakenet}.
By comparing \textsl{FE~w/o~GDP} with the default PointAttN, we can find that the average CD can be 10.6\% lower (6.63 versus 7.42) by utilizing GDPs, which demonstrates that perceiving geometric details is crucial for feature extraction.
The comparison between \textsl{FE~w/o~SFA} and the default PointAttN demonstrates that SFA also plays a role to improve the performance by enhancing the global information. 
By comparing \textsl{FE+FoldingNet} with the baseline FoldingNet \cite{foldingNet}, we can find that \textsl{FE+FoldingNet} reduces the average CD by 10.79, which is 56.6\% lower than FoldingNet \cite{foldingNet} (8.28 versus 19.07). The improvement by such a large margin demonstrates that our Feature Extractor can be uniquely adopted as a better backbone for feature extraction in point cloud analysis tasks.
By comparing \textsl{FE+SG+SPD} with the baseline SnowflakeNet \cite{snowflakenet}, we can find that the proposed Feature Extractor and Seed Generator outperform the default SnowflakeNet by achieving 3.8\% lower average CD (7.31 versus 7.60).
Compared with SA-Net \cite{SA-Net} that also adopts PointNet++ \cite{pointnet++} as backbone for feature extraction, \textsl{PN+SG+PG} reduces the average CD in a large margin by 35.6\% lower (7.23 versus 11.22), which demonstrates the capability of our coarse-to-fine decoding modules designed upon SFAs.
Moreover, by comparing all the five variations with the default PointAttN, each part shows its contribution to the overall performance improvement, which demonstrates the framework is neatly designed without useless part.

\section{Conclusions}
In this paper, we present a novel deep neural network, namely PointAttN, which has fully exploited the attention mechanism for point cloud completion. The main motivation of our work is to enlarge the receptive fields on processing 3D point clouds to better preserve their geometric structural and context information. Aiming to that, we leverage the cross-attention and self-attention mechanisms to perceive the local details and global context of 3D point clouds, by which the proposed framework eliminates the explicit local region partition like kNNs. Extensive comparisons and ablation studies are conducted to demonstrate the superiority of our proposed PointAttN, which outperforms the state-of-the-art methods on the Completion3D and PCN datasets.

\bibliographystyle{unsrt}  
\bibliography{references}

\end{document}